\documentclass[a4paper]{article}
\usepackage[textwidth=14.5cm]{geometry}
\usepackage{lipsum}

\usepackage{hyperref}       
\usepackage{url}            
\usepackage{booktabs}       
\usepackage{amsfonts}       

\usepackage{graphicx}
\usepackage[]{amssymb}
\usepackage[]{amsmath}
\usepackage{multirow}
\usepackage{stackengine}
\usepackage{scalerel}
\usepackage{authblk}
\usepackage{placeins}
\stackMath

\begin{document}
\title{Stochastic reconstruction of an oolitic limestone by generative adversarial networks.}

\author{Lukas Mosser \thanks{lukas.mosser15@imperial.ac.uk}}
\author{Olivier Dubrule \thanks{o.dubrule@imperial.ac.uk}}
\author{Martin J. Blunt \thanks{m.blunt@imperial.ac.uk}}
\affil{\small Department of Earth Science and Engineering, \\ Imperial College London}

\maketitle
\begin{abstract}
Stochastic image reconstruction is a key part of modern digital rock physics and materials analysis that aims to create 
numerous representative samples of material micro-structures for upscaling, numerical computation of effective properties and uncertainty quantification. 
We present a method of three-dimensional stochastic image reconstruction based on generative adversarial neural networks (GANs). 
GANs represent a framework of unsupervised learning methods that require no a priori inference of the probability distribution associated with the training data. Using a fully convolutional neural network allows fast sampling of large volumetric images. 
We apply a GAN based workflow of network training and image generation to an oolitic Ketton limestone micro-CT dataset. Minkowski functionals, effective permeability as well as velocity distributions of simulated flow within the acquired images are compared with the synthetic reconstructions generated by the deep neural network. While our results show that GANs allow a fast and accurate reconstruction of the evaluated image dataset, we address a number of open questions and challenges involved in the evaluation of generative network based methods.
\end{abstract}
\section{Introduction}
\label{sec:introduction}
The micro-structural characteristics of porous media play an important role in the understanding of numerous scientific and engineering applications such as the recovery of hydrocarbons from subsurface reservoirs \cite{blunt2013pore}, sequestration of $CO_2$ \cite{singh2017dynamics} or the design of new batteries \cite{SIDDIQUE2012437}.
Modern micro-computer tomographic (micro-CT) methods have enabled the acquisition of high-resolution three-dimensional images at the scale of individual pores. Increased resolution comes at the cost of longer image acquisition time and limited sample size. Individual samples allow numerical and experimental assessment of the effective properties of the porous media, but give no insight into the variance of key micro-structural properties. Therefore, an efficient method to generate representative volumetric models of porous media that allow the assessment of the effective properties is required. The generated images serve as an input to a digital rock physics workflow to represent the computational domain for numerical estimation of key physical properties \cite{BERG2017131}.

Typically, a statistical model allows for incorporation of measured data or a priori knowledge of structural elements.  The simulated annealing algorithm allows high-quality three-dimensional reconstruction and incorporation of numerous statistical descriptors of the porous medium \cite{yeong1998,jiao2008,pant2016stochastic}. Covariance-based methods such as truncated Gaussian simulation, allow the three-dimensional reconstruction of micro structures given two-point correlation functions. Three other stochastic approaches have been used in the past to reconstruct three-dimensional porous media outlined below.

Object based methods describe the material domain by imposing a spatial point process on a set of geometric bodies. The so-called Boolean model, considers a union of randomly placed overlapping objects of random shape, typically spheres \cite{matheron1975random,chiu2013stochastic}. Object based methods are a flexible framework \cite{torquato2013random} that may also allow interaction of particles to be incorporated and have successfully been used to describe complex and heterogeneous materials.  

Process models reconstruct the pore and grain structure of materials by mimicking their natural generating processes. Bakke and {\O}ren \cite{Bakke2003} have created reconstructions of sandstones by reproducing the natural processes of sedimentation, compaction and diagenesis. 

Where images of the materials micro structure can be obtained, by direct imaging or modeling, training image based methods allow stochastic reconstruction of the material domain. Training image based algorithms such as multiple-point statistics (MPS) \cite{mariethoz2014multiple}, direct sampling \cite{mariethoz2010direct} or image quilting have been successfully used in three-dimensional reconstruction of porous media \cite{Okabe2004,Okabe2007,mahmud2014}. 

This contribution presents a training image based method of image reconstruction using a class of deep generative methods called generative adversarial networks (GANs) first introduced by \cite{Goodfellow}. Recently, Mosser et. al \cite{Mosser17} have shown that GANs allow the reconstruction of three-dimensional porous media based on segmented volumetric images. Their study applied GANs to three segmented images of natural porous media. They showed that GANs represent a computationally efficient method that allow fast generation of large volumetric images that capture the statistical and morphological features, as well as the effective permeability. 

We expand on the work of Mosser et. al \cite{Mosser17} and investigate the ability of generative adversarial networks to create stochastic reconstructions of an unsegmented micro-CT scan of a larger oolitic Ketton limestone sample. Using the unsegmented gray-scale image dataset allows us to evaluate the four Minkowski functionals for the three-dimensional datasets as a function of the gray-level threshold. In addition to the numerical evaluation of permeability as shown by Mosser et al.\cite{Mosser17}, we compare velocity distributions of the original porous medium and samples obtained from the GAN.

\section{Generative Adversarial Networks}
\label{sec:generative_adversarial_networks}

Generative adversarial networks have been developed in the framework of deep generative methods as a method to generate samples from arbitrary probability distributions \cite{goodfellow2014,goodfellow2016}. GANs do not impose any a priori model on the probability density and are therefore also referred to as an implicit method. Without the need to specify an explicit model, GANs provide efficient sampling methods for high dimensional and intractable density functions. 

In the case of CT images of porous media we can define an image $x$ to be a sample of a real, unknown probability density function (pdf) of images $p_{data}$ of which we have acquired a number of samples which serve as training images. The training set is comprised of $64^3$ voxel sub-domains of the original micro-CT image. Sub-domains are extracted without any overlap and each training image represents an independent part of the originally acquired dataset.

GANs consist of two functions: a generator, whose role it is to generate samples of the unknown density $p_{data}(\mathbf{x})$ and a discriminator function that tries to distinguish between samples from the training set and synthetic images created by the generator. The generator $G$ is defined by its parameters $\mathbf{\theta}$ and performs a mapping from a noise prior $\mathbf{z}$ to the image domain:

\begin{equation}
\mathbf{z} \sim \mathcal{N}(0, 1)^{d \times 1 \times 1 \times 1}
\label{equ:noise_prior}
\end{equation}
\begin{equation}
 G_{\mathbf{\theta}}: \mathbf{z} \rightarrow \mathbb{R}^{1 \times 64 \times 64 \times 64}
\label{equ:generator_mapping}
\end{equation}
where $d$ is the dimensionality of the noise prior.

The discriminator $D_{\mathbf{\omega}}(\mathbf{x})$ assigns a probability to an image $x$ being a sample of the true data distribution $p_{data}$:
\begin{equation}
\label{equ:discrminator_mapping}
 D_{\mathbf{\omega}}: \mathbb{R}^{1 \times 64 \times 64 \times 64} \rightarrow [0, 1]
\end{equation}
where values close to 1 represent a high probability of being a sample of $\mathbf{x} \sim p_{data}(\mathbf{x})$.

We represent both the generator $G_{\mathbf{\theta}}(\mathbf{z})$ and the discriminator $D_{\mathbf{\omega}}(\mathbf{x})$ by differentiable neural networks with parameters $\mathbf{\theta}$ and $\mathbf{\omega}$ respectively. This allows us to use backpropagation combined with mini-batch gradient descent to optimize the generator and discriminator according to the functional:
\begin{equation}
\label{equ:minmax}
\min_{\mathbf{\theta}} \max_{\mathbf{\omega}}\{\mathbb{E}_{\mathbf{x}\sim p_{data}}[log \ D_{\mathbf{\omega}}(\mathbf{x})] + \mathbb{E}_{\mathbf{x}\sim p_{\mathbf{z}}}[log \ 1-D_{\mathbf{\omega}}(G_{\mathbf{\theta}}(\mathbf{z}))]\}
\end{equation}

The optimization criterion of the generator and discriminator (Eq.~\ref{equ:minmax}) is solved sequentially in a two-step procedure. We first train the discriminator to maximize its ability to distinguish real from fake samples. This is done in a supervised manner by training the discriminator on known real samples (Label 1) and samples created by the generator (Label 0). The misclassification error is back-propagated while keeping the parameters of the generator constant.

In a second step we train the generator to maximize its ability to "fool" the discriminator into misclassifying the images provided by the generator as real images. The parameters of the generator are then modified by applying stochastic gradient descent while keeping the parameters of the discriminator fixed. 

Training of these networks is often challenging due to the competing objective functions of the generator and discriminator. Recently, new objective functions and training heuristics have greatly improved the training process of GANs \cite{arjovsky2017wasserstein,2017arXiv170310717B}.

Training GANs follows a different scheme from other stochastic reconstruction methods (Sec.~\ref{sec:introduction}). There are two phases in GAN based reconstruction: training and generation. Training is expensive, requiring modern graphics processing units (GPU) and for three-dimensional datasets large GPU memory due to the three-dimensional nature of the filter maps and images. Parallelization of the training process across numerous GPUs reduces time for training the network. Nevertheless, finding a set of hyper-parameters that is a network architecture (number of filters, types, order of layers and activation functions) that leads to the desired quality can require significant trial and error. 

The second phase of GAN based reconstruction, the generation of individual samples, is extremely fast. All operations in the generator network can be represented as matrix-vector operations which are executed efficiently on modern GPU systems and take on the order of seconds for modern CPUs. 
\begin{figure}
  \includegraphics[keepaspectratio=True, width=\textwidth]{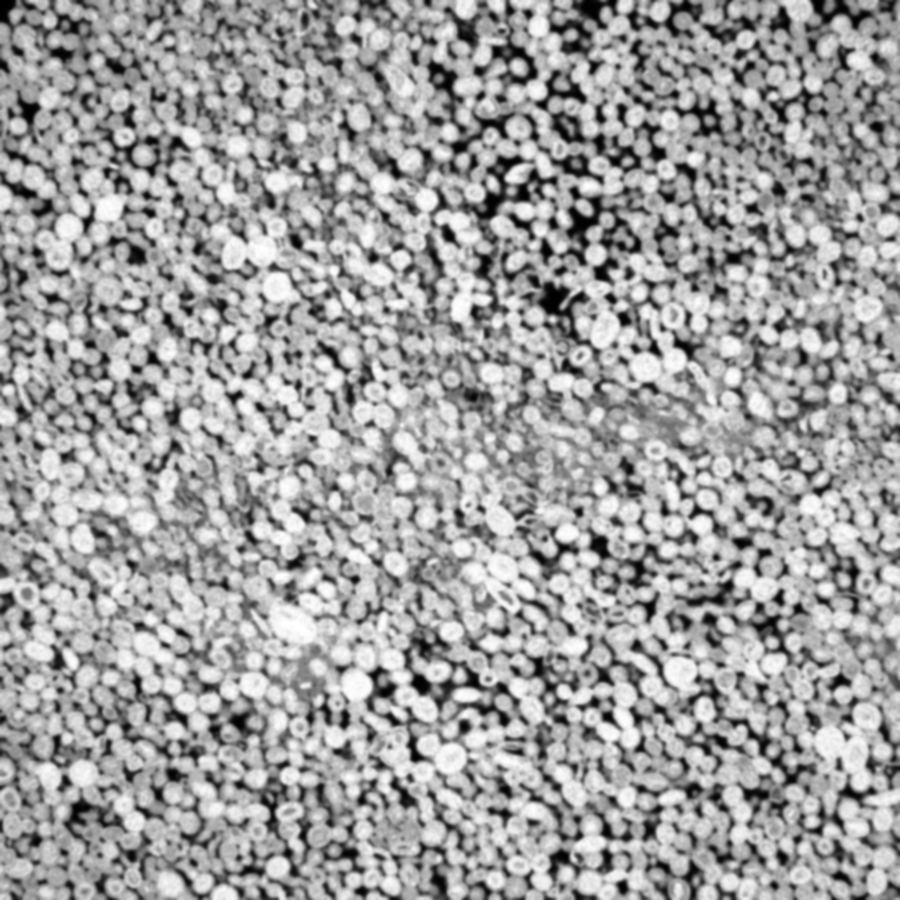}
\caption{Two-dimensional cross-section of the three-dimensional micro-CT image of the studied oolitic Ketton limestone sample. The image was acquired with a voxel size of 27.8 $\mu m$. Histogram equalization was applied to the image prior to its use as a training image.}
\label{fig:ketton_image}
\end{figure}
\FloatBarrier
\section{Dataset}
\label{sec:dataset}

The sample used in this study is an oolitic limestone of Jurassic age (169\textemdash176 million years). The spherical to ellipsoidal grains consist of 99.1\% calcite and 0.9\% quartz \cite{menke2017dynamic}. Inter and intra-granular porosity can be observed, as well as significant amounts of unresolved sub-resolution micro-porosity. This is characterized by the various shades of gray in individual grains, where the interaction of sub-resolution porosity with x-rays penetrating the sample during imaging leads to an increase in intermediate gray-level values (Fig.~\ref{fig:ketton_image}). The sample was imaged using a Zeiss XRM 510 with a voxel size of 27.8 $\mu m$. The size of the image domain after resampling to 8bit resolution is $900^3$ voxels. We subdivide the original image into a training set of non-overlapping 5832 images at a size of $64^3$ voxels. We define a sequential randomized pass over the full training set as an epoch. Evaluation of the effective properties is performed at larger image sizes than the training images to judge whether the GAN is able to generalize to larger domains. To evaluate the reconstruction quality of the GAN model we randomly extract 64 images at a size of $200^3$ voxels with no overlap from the original training image (Fig.~\ref{fig:ketton_image}) which we refer to as the validation set. A synthetic validation set was created by sampling 64 images at a size of $200^3$ voxels from the trained GAN model. To perform numerical computation of the effective permeability as well as measure the two-point correlation function, all images of the synthetic and original Ketton validation set were segmented using Otsu thresholding \cite{otsu1975threshold}. Minkowski functionals were evaluated for the unsegmented validation sets.

\subsection{Neural Network Architecture and Training}
\label{sec:neural_network}
Following the work of Radford et. al~\cite{Radford2016} we use a fully convolutional version of the DCGAN architecture. We replace the fully connected input layer of the generator in DCGAN by a transposed convolution. The fully convolutional nature of the generator allows us to create images of arbitrary size by providing latent vectors with larger spatial dimensionality e.g. $\mathbf{z} \sim \mathcal{N}(0, 1)^{d \times m \times n \times o}$. During training $m$, $n$ and $o$ are of size one, which results in an image of $64^3$ voxels. For image generation $m$, $n$ and $o$ may be of any integer size.

\begin{figure}
  \includegraphics[keepaspectratio=True, width=\textwidth]{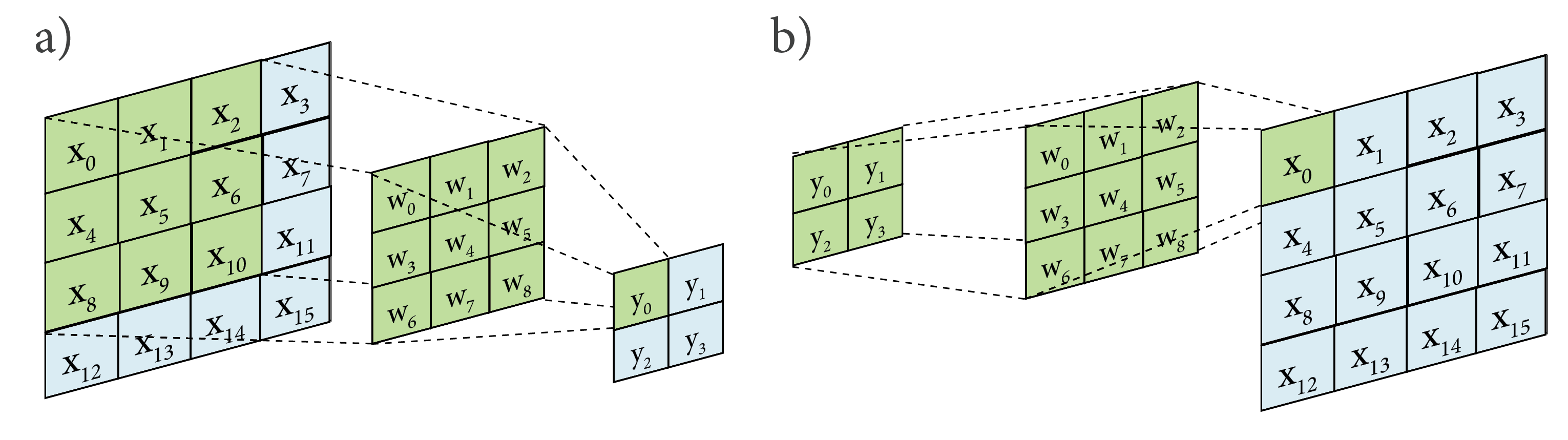}
\caption{Example of a discrete convolution (a) and equivalent transposed convolution operation (b) for a $3 \times 3$ filter kernel size applied to a $4\times 4$ feature map. The active regions to compute the output value are shaded green.}
\label{fig:convolution_operation}
\end{figure}

In Figure~{\ref{fig:convolution_operation}} we show an example of a convolution and transposed convolution operation for the two-dimensional case. The convolution is performed by sliding a filter kernel $w_i$ (Eq.~\ref{weights_unrolled}) over the input feature map $x_i$ (Eq.~\ref{output}) \cite{dumoulin2016guide}. We rewrite this as an efficient matrix vector operation (Eq.~\ref{convolution}) by unrolling the discrete convolution:~\begin{equation}
\mathbf{W}=\begin{pmatrix}
 w_0 & w_1 & w_2 & 0 & w_3 & w_4 & w_5 & 0 & w_6 & w_7 & w_8 & 0 & 0 & 0 & 0 & 0 \\
 0 & w_0 & w_1 & w_2 & 0 & w_3 & w_4 & w_5 & 0 & w_6 & w_7 & w_8 & 0 & 0 & 0 & 0 \\
 0 & 0 & 0 & 0 & w_0 & w_1 & w_2 & 0 & w_3 & w_4 & w_5 & 0 & w_6 & w_7 & w_8 & 0 \\
 0 & 0 & 0 & 0 & 0 & w_0 & w_1 & w_2 & 0 & w_3 & w_4 & w_5 & 0 & w_6 & w_7 & w_8
 \end{pmatrix} \ [4 \times 16] \label{weights_unrolled}
\end{equation}

\noindent~The input image $\mathbf{x}$, in this case a single-channel $4\times4$ image, and the output $\mathbf{y}$ are represented as one dimensional vectors:~\begin{equation}
 \mathbf{x} \ [16 \times 1], \ \mathbf{y} \ [4 \times 1] \label{output} 
\end{equation}
\noindent~This allows us to perform the discrete convolution:
\begin{equation}
 \mathbf{W} * \mathbf{x} = \mathbf{y} \label{convolution}
\end{equation}
\noindent~and we can define the transpose operation:
\begin{equation}
 \mathbf{W}^T * \mathbf{y'} = \mathbf{x'} \label{convolution_transposed}
\end{equation}
\noindent~where $\mathbf{y}'$ and $\mathbf{x}'$ are defined according to Eq.~\ref{output}.
For each convolutional layer of the network, the input features are convolved with a number of independent filter kernels $\mathbf{W}$.
\FloatBarrier
\newpage
\begin{figure}
  \includegraphics[keepaspectratio=True, width=\textwidth]{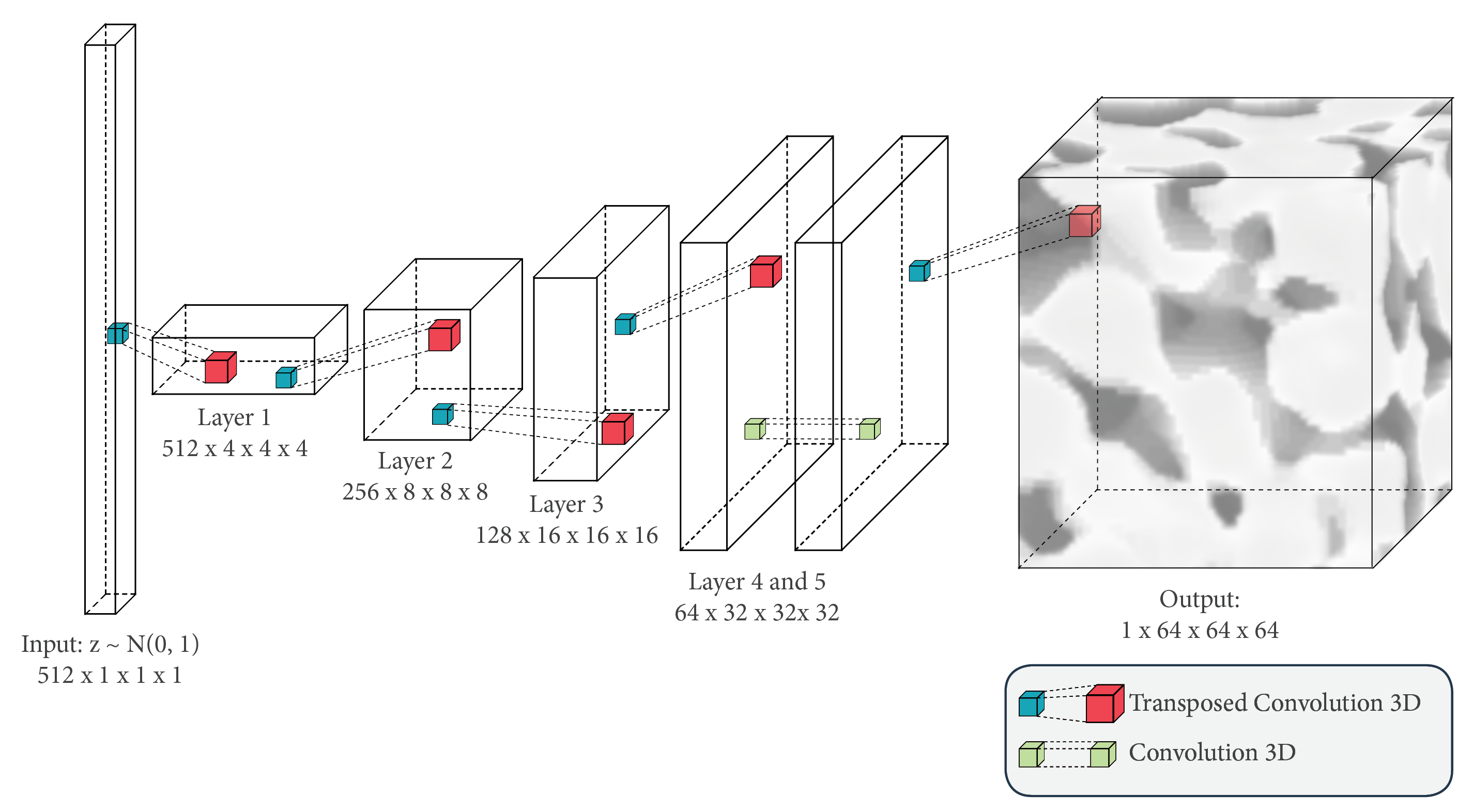}
\caption{Architecture of the neural network used to represent the generator function $G_{\theta}(\mathbf{z})$. The latent vector $\mathbf{z}$ is passed through a fully convolutional feed-forward neural network. Transposed convolution operations upsample the image in each layer. A single convolutional layer is introduced prior to the final network layer to reduce artifacts due to upsampling using transposed convolution.}
\label{fig:neural_network_architecture}
\end{figure}
\FloatBarrier

\begin{figure}
  \includegraphics[keepaspectratio=True, width=\textwidth]{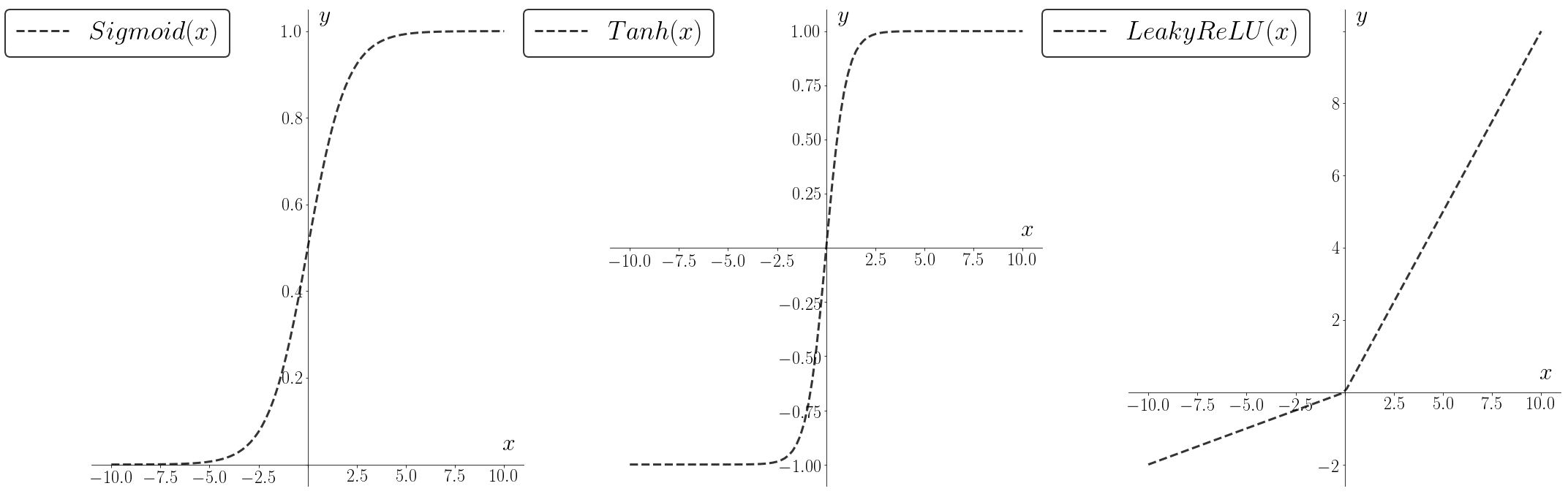}
\caption{Activation functions used in the generator and discriminator networks.}
\label{fig:activation_functions}
\end{figure}

\noindent~The generator consists of a series of three-dimensional transposed convolutions. In each layer, the number of weight kernels is reduced by a factor of $\frac{1}{2}$. Before the final transposed convolution we add an additional convolutional layer (Fig.~\ref{fig:neural_network_architecture}). Each layer in the network except the last is followed by a batch normalization \cite{2015arXiv150203167I} and a Leaky Rectified Linear Unit (LeakyReLU) activation function. The final transposed convolution in the generator is followed by a hyperbolic tangent activation function (Tanh). A representation of each activation function used in the network is shown in Fig.~\ref{fig:activation_functions}.

We represent the discriminator as a convolutional classification network with binary output using as input the real samples of the $64^3$ voxel training set (Label 1) and synthetic realizations of same size created by the generator (Label 0). Each layer in the network consists of a three-dimensional convolution operations followed by batch normalization and a LeakyReLU activation function. The final convolutional layer outputs a single value between 0 and 1 (Sigmoid activation) which corresponds to the probability that the input image belongs to the original training set or in other words that it is a real image.

We distinguish two sets of parameters for training: the set of weights of a network comprises the adjustable parameters of the filter kernels for convolutional and neurons for linear network layers. The so-called hyper-parameters define the network architecture and training scheme i.e. the number of filters per layer, the number of filters or learning rates. A different set of hyper-parameters defines different networks with their own weights (parameters) which are adapted using a mini-batch gradient descent method at training time. In total 8 models have been trained on the Ketton image dataset. The main hyper-parameters that were varied for each model are the number of filters in the generator and discriminator, $N_{GF}$ and $N_{DF}$ respectively, as well as the number of convolutional layers before the final transposed convolution in the generator. The dimensionality of the latent vector $\mathbf{z}$ was kept constant at a size of $512\times 1 \times 1 \times 1$. Learning was performed by stochastic gradient descent using the ADAM optimizer with a momentum of $\beta_1=0.5$, $\beta_2=0.999$ and a constant learning rate of $2 \times 10^{-4}$. Network training was performed on eight NVIDIA K40 GPUs.

To train the pair of networks $G_{\mathbf{\theta}}(\mathbf{z})$ and $D_{\mathbf{\omega}}(\mathbf{x})$ we make use of two heuristic stabilization methods. First, Gaussian noise $(\mu=0, \sigma=0.1)$ is added to the input of the discriminator which is annealed linearly over the first 300 epochs of training \cite{2016arXiv161004490K}. We perform three steps of training the discriminator per generator step. To stabilize GAN training further, we perform so-called label switching on every third discriminator training step. This heuristic stabilization method is performed by training the discriminator for one step with switched labels; a real image is expected to be labeled as false and generated images as real. Among the eight models tested, the network architecture of the model with the smallest error is presented in Table~\ref{tab:gan_architecture}. The presented model has hyper-parameters of $N_{DF}=N_{GF}=64$. Training was stopped after 170 epochs i.e. full iterations of the training set of images.  The generator consists of $27.9\times 10^6$ adjustable parameters and $11.0\times 10^6$ parameters for the discriminator. Visual inspection of the generated images and empirical computation of morphological and statistical properties were used as a measure for reconstruction performance at each iteration.

\begin{figure}
  \includegraphics[keepaspectratio=True, width=\textwidth]{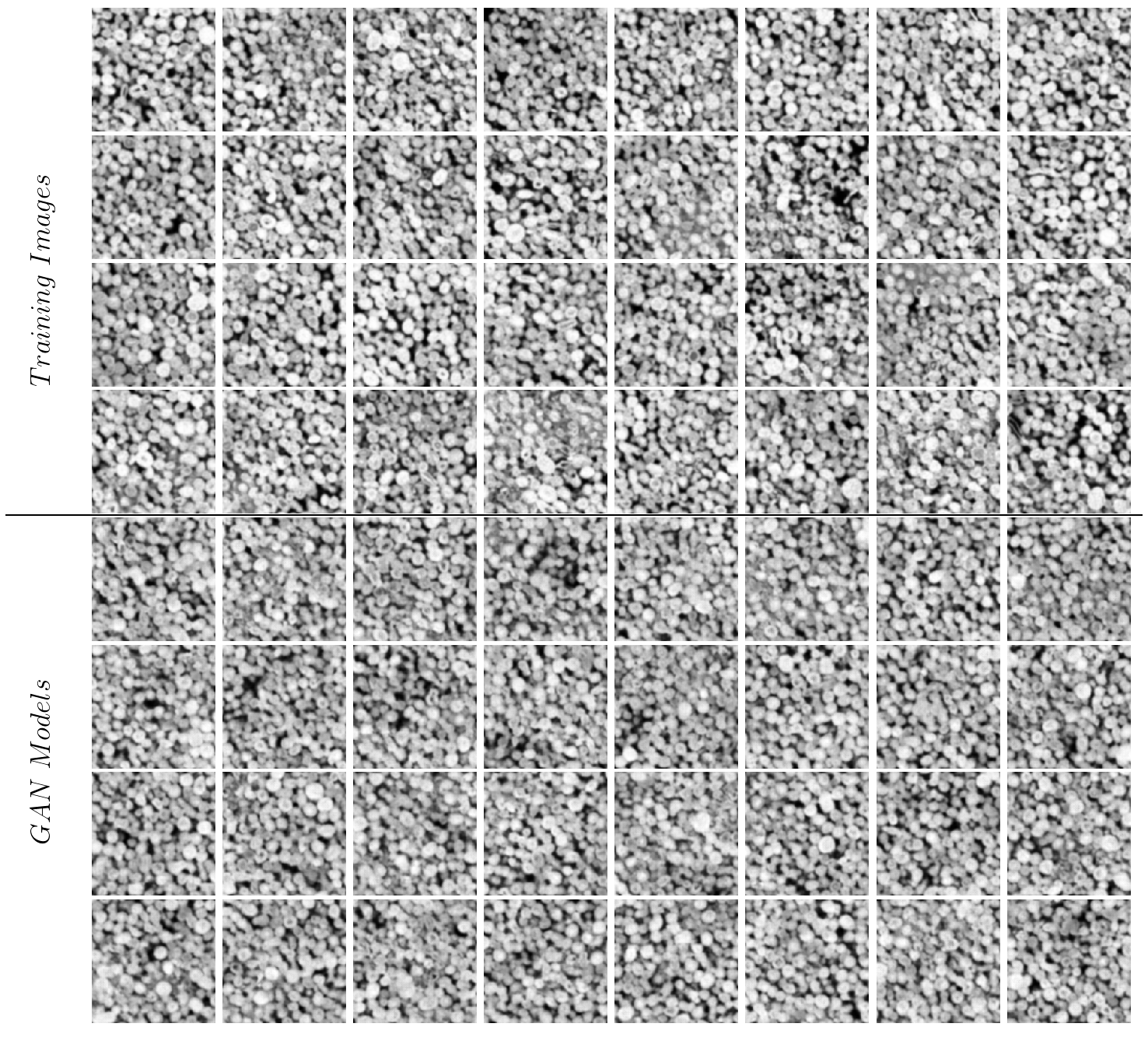}
\caption{Cross-sections of the $200^3$ voxel sub-domains of the Ketton micro-CT image (top) and synthetic realizations obtained from the trained generator of the generative network (bottom).}
\label{fig:comparison_data}
\end{figure}

\begin{table}
\centering
\caption{Architecture of the generator and discriminator networks. The generator is a fully convolutional version of a DCGAN \cite{Radford2016} with one additional convolution layer prior to the final transposed convolution. LeakyReLU activation functions were used for all layers except the last.}
\label{tab:gan_architecture} 
\begin{tabular}{cccccccc}
\smallskip \\
\multicolumn{1}{l}{} & \multicolumn{7}{c}{Generator}    \\
\hline\smallskip \\
Layer                & Type          & Filters & Kernel                & Stride & Padding & Batchnorm & Activation \\
\hline\smallskip \\
1                    & ConvTransp3D& 512     & $4 \times 4 \times 4$ &  1     & 0       & Yes       & LeakyReLU  \\
2                    & ConvTransp3D&  256       &  $4 \times 4 \times 4$ &  2      &    1     &  Yes         &   LeakyReLU \\ 
3                    & ConvTransp3D&  128       &  $4 \times 4 \times 4$ &  2      &    1     &  Yes         &   LeakyReLU \\ 
4                    & ConvTransp3D&  64       &  $4 \times 4 \times 4$ &  2      &    1     &  Yes         &   LeakyReLU \\ 
5                    & Conv3D&  64       &  $3 \times 3 \times 3$ &  1      &    1     &  Yes         &   LeakyReLU \\ 
6                    & ConvTransp3D&  1       &  $4 \times 4 \times 4$ &  2      &    1     &  No         &   Tanh \\ 
\hline\smallskip \\
\multicolumn{1}{l}{} & \multicolumn{7}{c}{Discriminator}    \\
\hline\smallskip \\
Layer                & Type          & Filters & Kernel                & Stride & Padding & Batchnorm & Activation \\
\hline\smallskip \\
1                    & Conv3D&64     & $4 \times 4 \times 4$ & 2     & 1      & No       & LeakyReLU  \\
2                    & Conv3D&128     & $4 \times 4 \times 4$ &  2     & 1       & Yes       & LeakyReLU  \\
3                    & Conv3D& 256     & $4 \times 4 \times 4$ &  2     & 1       & Yes       & LeakyReLU  \\
4                    & Conv3D& 512     & $4 \times 4 \times 4$ &  2     & 1       & Yes       & LeakyReLU  \\
5                    & Conv3D& 1     & $4 \times 4 \times 4$ &  1     & 0       & No       & Sigmoid  \\
\end{tabular}
\end{table}
After training, the generator was used to create 64 reconstructions at a size of $200^3$ voxels by sampling from the noise prior $\mathbf{z}$ (Eq.~\ref{equ:noise_prior}) and performing the mapping from the latent space to the image space (Eq.~\ref{equ:generator_mapping}). Figure~\ref{fig:comparison_data} shows slices through 32 non-overlapping sub-domains of the Ketton validation set and slices through 32 synthetic validation samples generated by the GAN model. The shown samples represent a random set of the generator output and were not selected by hand for their visual or statistical quality. The following sections present the a posteriori calculations of statistical, morphological and effective properties of these 64 synthetic validation images in comparison to the extracted validation set of the original Ketton image~(Fig.~\ref{fig:comparison_data}).

\subsection{Two-Point Probability Functions}
\label{sec:two_point_probability}
The two-point probability functions $S_2(\mathbf{r})$ allow the first and second order moments of a micro-structure to be characterized. We define the isotropic non-centered two-point probability function $S_2(\mathbf{r})$ as the probability that two arbitrary points separated by a distance $\|\mathbf{r}\|$ are located in the same phase i.e. grain or void phase of the micro-structure. While $S_2(\mathbf{r})$ may be defined for both phases of a porous medium, we compute the two-point probability function with respect to the pore phase only.

\begin{equation}
S_2(\mathbf{r})=\mathbf{P}(\mathbf{x} \in P, \mathbf{x}+\mathbf{r} \in P) \ for \ \mathbf{x}, \mathbf{r} \in \mathbb{R}^d \label{equ:covariance}
\end{equation}

\noindent~$S_2(0)$ is equal to the porosity of the porous medium. Stabilization of $S_2(\mathbf{r})$ occurs around a value of $\phi^2$ as the distance tends towards infinity. In addition, the specific surface area $S_V$ can be determined from the slope of the two-point probability function at the origin $S_V = -4S_2'(0)$ \cite{berryman1987relationship}.

\begin{figure}
  \includegraphics[keepaspectratio=True, width=\textwidth]{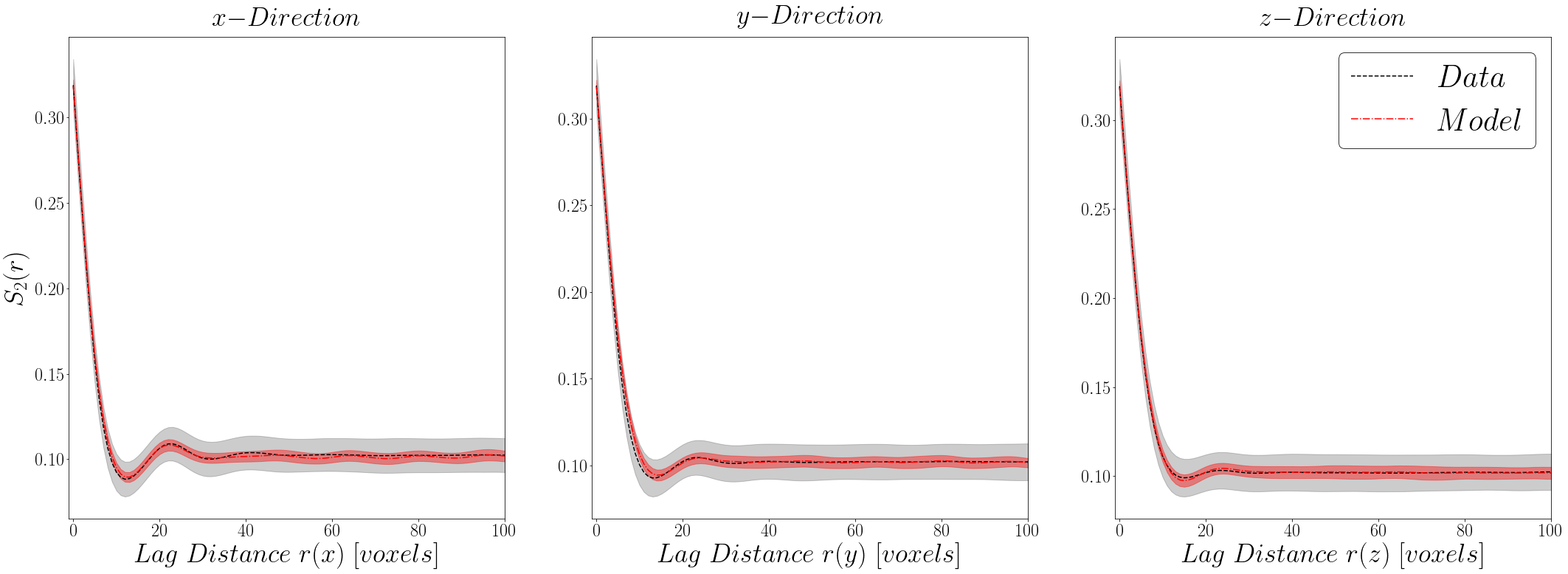}
\caption{Comparison of the two-point probability function $S_2(\mathbf{r})$ measured along the Cartesian axes for Ketton image sub-domains and GAN generated realizations. $S_2(\mathbf{r})$ was measured on images after thresholding using Otsu's method. Grey and red shaded areas respectively show the variation around the average behavior ($\mu \pm \sigma$) of 64 images of the Ketton image and GAN generated validation set.}
\label{fig:directional_s2}
\end{figure}

We calculate $S_2(\mathbf{r})$ numerically using the lattice point algorithm described by Jiao et. al \cite{jiao2008}. Figure~\ref{fig:directional_s2} shows the directional two-point probability function for 64 $200^3$ voxel sub-domains of the original Ketton validation set (gray) and the GAN generated realizations (red). Our findings show that the 64 GAN-generated realizations lie within the standard deviation of the experimental $S_2(\mathbf{r})$ computed for the 64 original Ketton images. 

Due to the ellipsoidal nature of the grains found in the Ketton limestone, a significant oscillation can be observed in all three orthogonal directions. This "hole-effect" is characteristic of periodic media \cite{torquato1985characterisation} and is a feature that should be reproduced by reconstructions of the original training images. The hole-effect found in the training image dataset is reproduced by the samples generated by the GAN model, indicating the preservation of periodic features in the pore micro-structure of the synthetic images.

\begin{figure}\centering
  \includegraphics[keepaspectratio=True, width=0.5\textwidth]{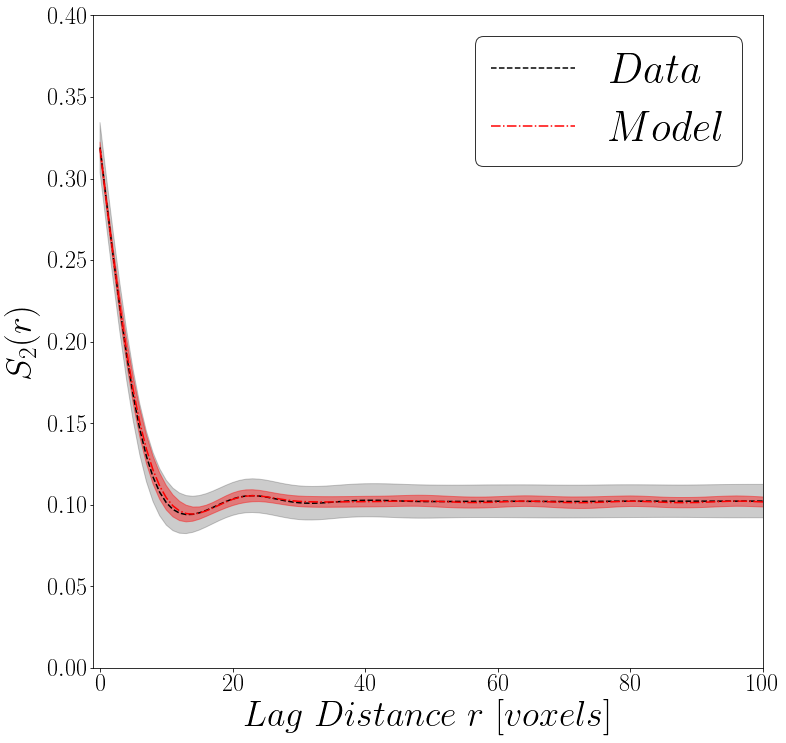}
\caption{Radial average of the average two-point probability function $S_2(\mathbf{r}))$ for 64 dataset sub-domains and GAN generated images. Excellent agreement of the average behavior can be observed (dashed line) whereas a lower variation around the mean behavior can be observed for the GAN generated images.}
\label{fig:radial_average_s2}
\end{figure}

Good agreement between the real and synthetic micro-structures can be observed for the radial averaged two-point probability function (Fig.~\ref{fig:radial_average_s2}). For both the radial averaged and directional estimates of $S_2(\mathbf{r})$ a tight clustering around the mean can be observed, where the real porous medium shows a larger degree of variation around the mean. 

\subsection{Minkowski Functionals}
\label{minkowski_functionals}
To evaluate the ability of the trained GAN model to capture the morphological properties of the studied Ketton limestone, we compute four integral geometric properties that are closely related to the set of Minkowski functionals as a function of the image gray value. 

For any n-dimensional body we can define n+1 Minkowski functionals to characterize morphological descriptor of the grain-pore bodies structures \cite{mecke2000additivity}. The Minkowski functional of zeroth-order is equivalent to the porosity of a porous medium and defined as :
\begin{equation}
\phi = M_0 = \frac{V_{pore}}{V} \label{equ:porosity}
\end{equation}
where $V_{pore}$ corresponds to the pore volume and $V$ to the bulk volume of the porous medium.

We measure the specific surface area $S_V$ defined as an integral geometric relationship: 
\begin{equation}
S_V = \frac{M_1}{V} = \frac{1}{V}\int{dS}\label{equ:specific_surface_area_minkowski}
\end{equation}
where $M_1$ is the Minkowski functional of first order. In three dimensions, $M_1$ corresponds to the surface area of the pore-grain interface. Both $S_V$ and $\phi$ can be obtained by estimation of the two-point probability function $S_2(\mathbf{r})$ (Sec.~\ref{sec:two_point_probability}). The specific surface area $S_V$ has dimensions of $\frac{1}{length}$ and its inverse can be used to define a characteristic length scale of the porous medium. 

The Minkowski functional of order 2, the integral of mean curvature, $M_2$, can be related to the shape of the pore space due to its measure of the curvature of pore-grain interface. We use a bulk volume average of the specific surface area defined as:
\begin{equation}
\kappa_V = \frac{M_2}{V} = \frac{1}{2V}\int{(\frac{1}{r_1}+\frac{1}{r_2})dS}\label{equ:specific_integral_of_mean_curvature}
\end{equation}
\noindent where $r_1$ and $r_2$ are the principal radii of curvature of the pore-grain interface.

The Euler characteristic, $\xi_V$, is a measure of connectivity that is proportional to the dimensionless third order Minkowski functional $M_3$:
\begin{equation}
\chi_V = \frac{M_3}{4 \pi V} = \frac{1}{4 \pi V}\int{\frac{1}{r_1 r_2}dS}\label{equ:specific_euler_characteristic_minkowski}
\end{equation}

We evaluate these four image morphologic properties at each of the 256 gray-level values of the $200^3$ voxel Ketton image sub-domains and the GAN generated realizations. This allows us to describe the porous medium as a set of characteristic functions dependent on a global truncation value $\rho$ for each of the four Minkowski functionals \cite{schmahling2006statistical,minkowskiVogel}. To compute the four properties at each threshold level $\rho$ the publicly available micro-structure analysis software library Quantim was used \cite{quantim}. 

\begin{figure}\centering
  \includegraphics[keepaspectratio=True, width=\textwidth]{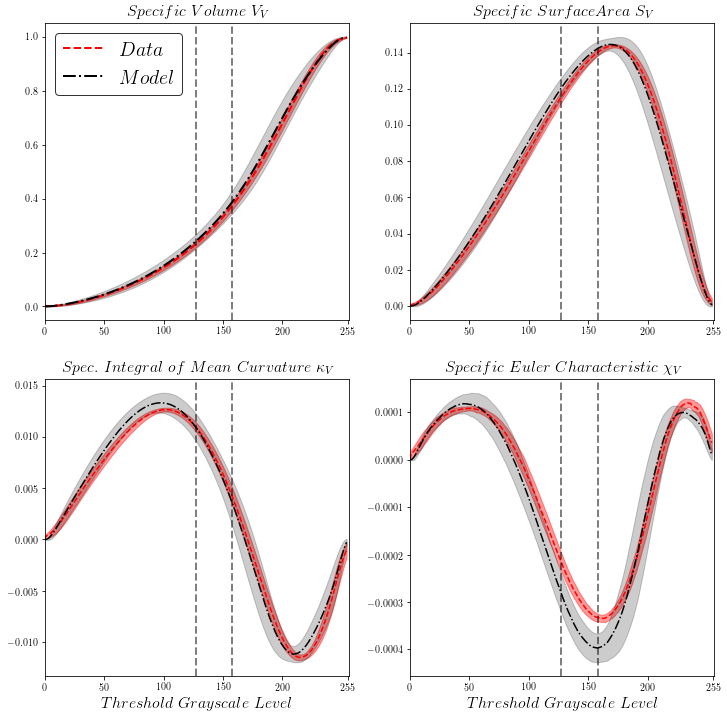}
\caption{Four Minkowski functionals as a function of the segmentation threshold. Shaded regions show variation of the properties around the mean $\mu \pm \sigma$. Vertical dashed lines show the region of segmentation thresholds obtained by applying Otsu's method.}
\label{fig:minkowski_functionals}
\end{figure}

Figure~\ref{fig:minkowski_functionals} compares these four estimated properties as a function of the image threshold value for the Ketton image (gray) and the samples generated by the GAN model (red). The shaded regions correspond to the variation around the mean $\mu \pm \sigma$ for both synthetic and real image datasets. The same 64 samples used in the evaluation of the two-point probability function have been used for this analysis. Additionally, the vertical dashed lines represent the range of the threshold values obtained by Otsu's method when applied to the individual images. This allows an estimate of the error region that is significant when introducing a thresholding method based on a global truncation value such as Otsu's method.

Our analysis of the GAN based models shows excellent agreement for the porosity $\phi(\rho)$, specific surface area $S_V(\rho)$ and integral of mean curvature $\kappa_V(\rho)$ as a function of the threshold value $\rho$. For these three properties a low error is introduced when applying global thresholding. The fourth property, the specific Euler characteristic, $\chi_V(\rho)$, shows an error of $20\%$ in the range of global thresholding values and in good agreement outside of this range of truncation values. This implies that care must be taken when segmenting an image --- real or generated --- to preserve the connectivity of the pore space.

\subsection{Permeability and Velocity Distributions}
\label{sec:results_permeability}
To validate GAN based model generation for uncertainty evaluation and numerical computations it is key that the generated samples capture the relevant physical properties of the porous media that the model was trained on. For the Ketton sample, the permeability and, moreover, the local velocity distributions represent the key properties of the porous medium \cite{menke2017dynamic}. 

To evaluate the ability of GAN-based models to capture the permeability and in-situ velocity distributions of the Ketton training images, we solve the Stokes equation on a segmented representation of each of the 64 Ketton sub-domains and 64 synthetic pore representations created by the GAN model. The segmented representations used to estimate the two-point probability functions were reused for this evaluation. A finite-difference based method adapted for binary representations of voxel-based pore representations was used to compute the effective permeability from the derived velocity field \cite{Mostaghimi2013}. The effective permeability was computed in the three Cartesian directions. 

\begin{subequations}
\label{equ:stokes}
\begin{eqnarray}
\nabla \cdot \mathbf{v}&=&0\label{equationa}
\\
\mu \nabla^{2} \mathbf{v}&=&\nabla p\label{equationb}
\end{eqnarray}
\end{subequations}
We present the resulting distribution of estimated permeability values as a function of the effective porosity:
\begin{equation}
\label{equ:effective_porosity}
\phi_{eff} = \frac{V_{flow}}{V}
\end{equation}
where $V_{flow}$ is the volume of the connected porosity.

\begin{figure*}
\centering
  \includegraphics[keepaspectratio=True, width=\textwidth]{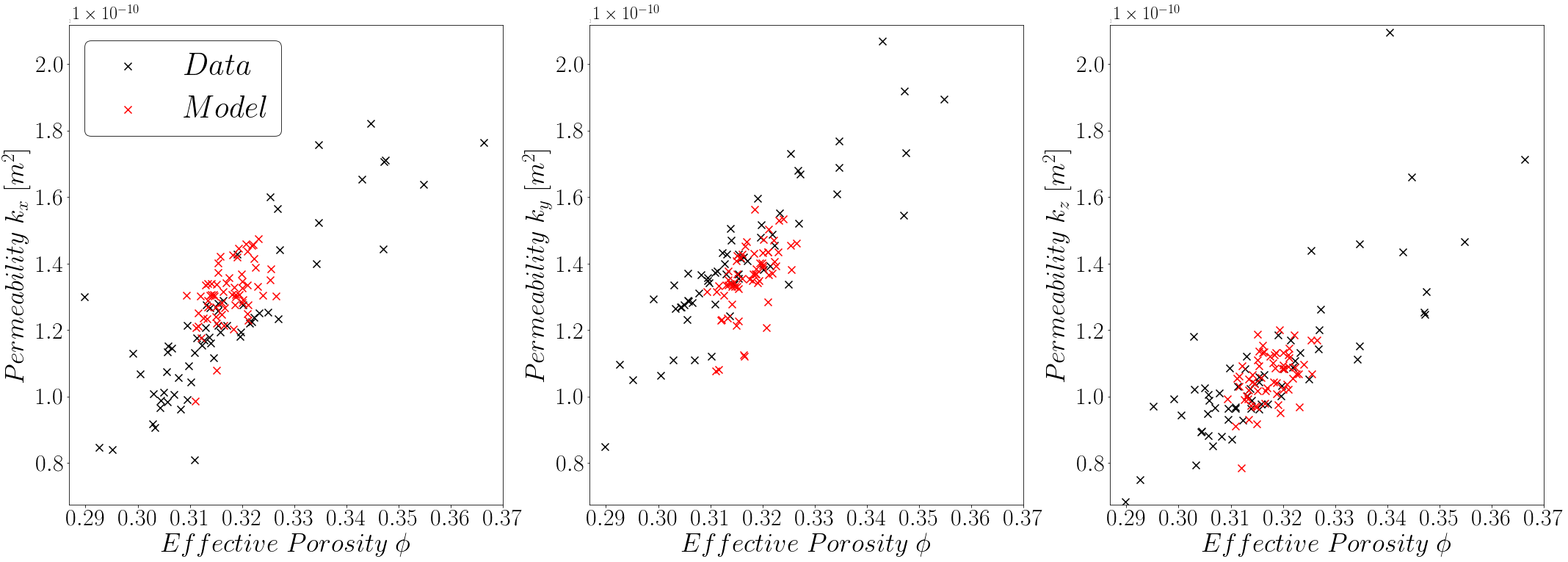}
\caption{Directional permeability computed on validation dataset (64 images at $200^3$ voxels) extracted from the original Ketton limestone micro-CT dataset and realizations obtained from the GAN model. Values of permeability obtained from the synthetic images are tightly clustered around the mean of the original dataset.}
\label{fig:directional_permeability}
\end{figure*}

\begin{figure}
\centering
  \includegraphics[keepaspectratio=True, width=0.5\textwidth]{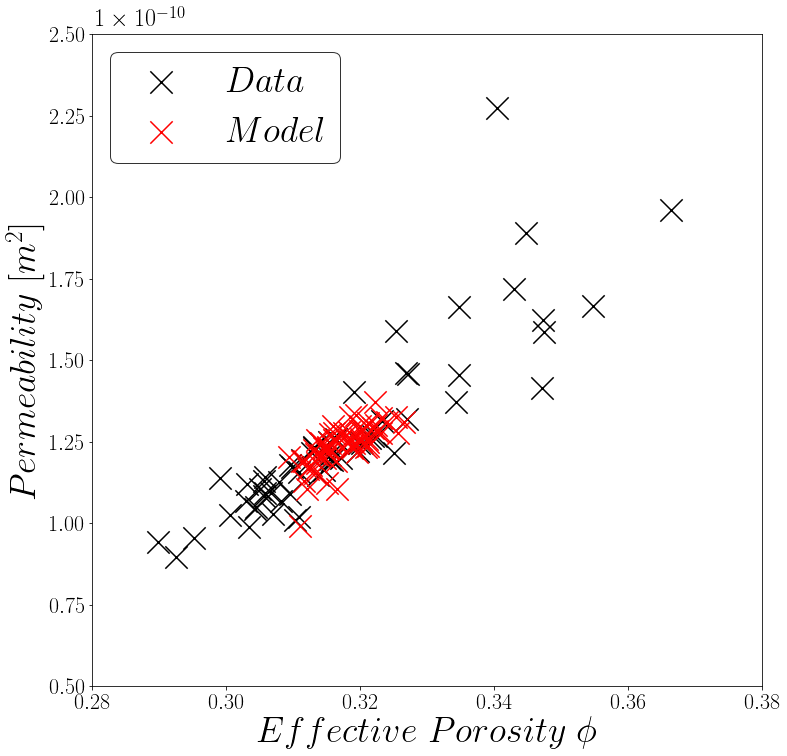}
\caption{Averaged permeability for the original image datasets and synthetic realizations obtained from the GAN model.}
\label{fig:average_permeability} 
\end{figure}

Our results show (Figs.~\ref{fig:directional_permeability} and \ref{fig:average_permeability}) that the GAN model generates stochastic reconstructions that capture the average permeability of the original training image at a scale of $200^3$ voxels, with the majority of samples closely centered around the average effective permeability of the Ketton subsets. 

The velocity distributions of the $2 \times 64$ numerical simulations were normalized by the average cell-centered velocity following the approach of Alhashmi et. al \cite{alhashmi2016impact} and a histogram with 256 logarithmically-spaced bins in a range from $1\times 10^{-4}$ to $1 \times 10^2$ for each simulation was obtained. 

\begin{figure}
  \includegraphics[keepaspectratio=True, width=\textwidth]{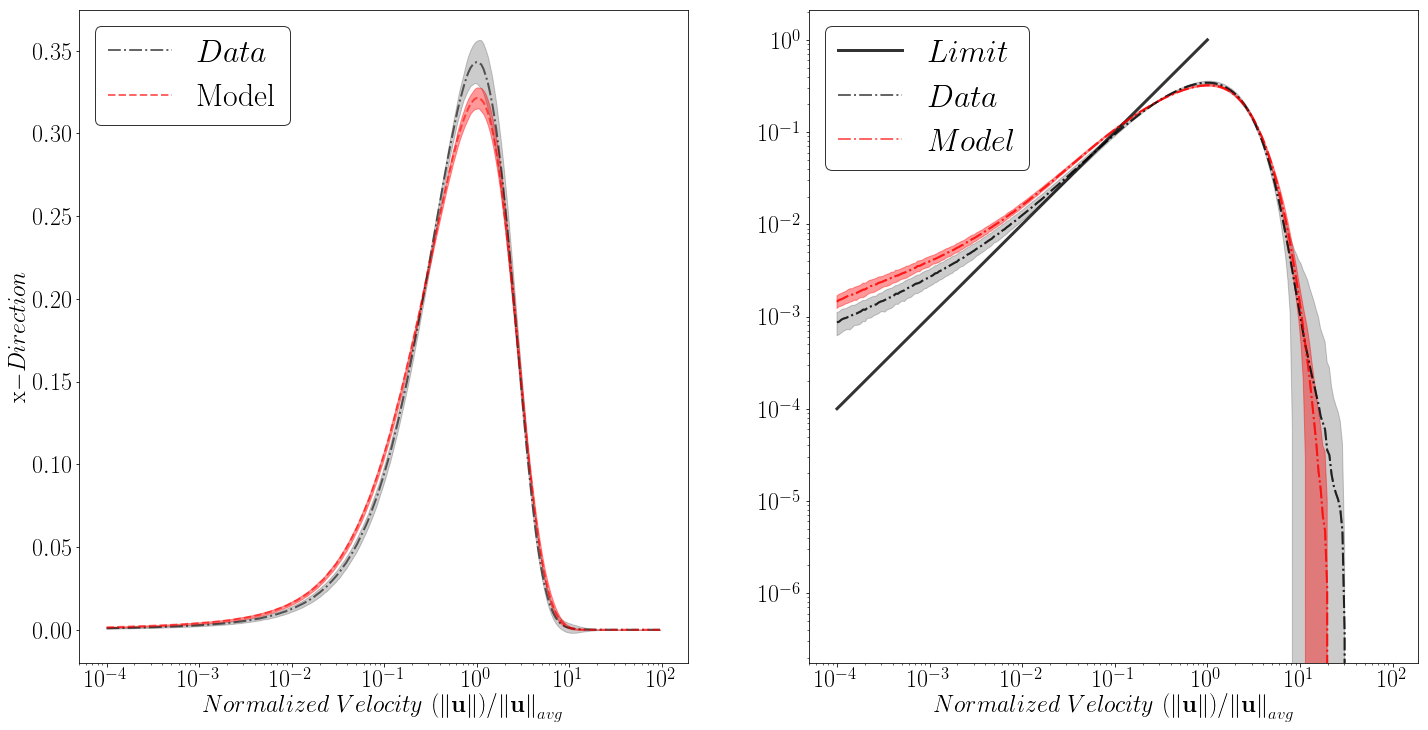}
\caption{Comparison of probability density functions of the magnitude of velocity extracted from the centers of voxels in the pore space divided by the average flow velocity plotted on semi-logarithmic (left) and double-logarithmic axes (right). The combination of 64 simulations of sub-domains obtained from the original dataset and 64 generated realizations of the GAN model are shown. Shaded regions highlight the variation around the mean of all simulations $\mu \pm \sigma$. The solid line shows the homogeneous limit velocity distribution for a single capillary tube.}
\label{fig:velocity_distributions}
\end{figure}

Figure~\ref{fig:velocity_distributions} shows the per bin arithmetic average of the bin frequencies and a bounding region of one standard deviation $\mu \pm \sigma$ as the shaded area. Due to the high range of velocities spanning six orders of magnitude,the x-axis is represented in logarithmic scaling. 

Visually, the distributions of the generated samples and Ketton sub-domains are nearly equivalent with minor deviations in the frequency of the very high and very low velocities. For the GAN model, low velocities are more abundant than in the original image whereas the opposite is true for high velocities.

To evaluate whether the velocity distributions obtained from numerical simulation of flow for the GAN generated images are statistically similar to distributions representative of the original image dataset we perform a two-sample Kolmogorov-Smirnov test. The null Hypothesis $H_0$ states that two samples are of the same underlying distribution. Define $D_{n,m}$ as:

\begin{equation}
\label{equ:dnm_statistic}
D_{n,m}=\sup_{x}|F_{1,n}(x)-F_{2,m}(x)|
\end{equation}
\noindent and the null hypothesis $H_0$ is rejected if
\begin{equation}
\label{equ:rejection_criterion}
D_{n,m} > c(\alpha){\sqrt{\frac{n+m}{nm}}}
\end{equation}
\noindent where $n$ and $m$ are the sample sizes respectively and $c(\alpha)=\sqrt{-\frac{1}{2}ln(\frac{\alpha}{2})}$. All experiments were performed at a significance level of $\alpha=0.05$ for the per-bin average velocity distributions presented in Fig.~\ref{fig:velocity_distributions} (dashed curves).

\begin{table}
\centering
\caption{Results of the two sample Kolmogorov-Smirnov test for equality of velocity distributions computed on the image dataset and generated realizations. The null hypothesis of distributional equality is to be accepted at a significance level of $\alpha=0.05$ for all three directional velocity distributions.}
\label{tab:1}       
\begin{tabular}{llll}
\hline\noalign{\smallskip}
\multicolumn{1}{l}{Direction} & \multicolumn{1}{l}{$D_{n,m}$}&  \multicolumn{1}{l}{$D_{0.05}$}   \\
\noalign{\smallskip}\hline\noalign{\smallskip}
x & 0.09 & \multirow{2}{*}{0.12} \\
y & 0.09 &    \\
z &	0.07 & 	  \\
\noalign{\smallskip}\hline
\end{tabular}
\end{table}

For all three directions the null hypothesis can be accepted at the 5\% significance level based on the $D_{0.05}$ statistic, giving evidence to the visual similarity between the velocity distributions of the real Ketton images and their synthetic counterparts~(Table~\ref{tab:1}).  

\section{Discussion}
\label{sec:discussion}
We have presented the results of training a generative adversarial network on a micro-CT image of the oolitic Ketton limestone. The image morphological properties were evaluated as a function of the image threshold level and it was shown that the generated images capture the textural features of the original training image. Two-point statistics and effective properties computed on segmented representations of the individual sub-domains have also shown excellent agreement between the realizations generated by the GAN model and subsets of the Ketton image. Nevertheless there remain a number of open questions that need to be addressed.

All effective properties, statistical and morphological properties have shown a tight bound around the average behavior of the training image. This indicates that there is less variation in the generated samples than in the training samples. This behavior can have a number of origins. 

The training images can be regarded as samples of the unknown multivariate pdf $p_{real}(\mathbf{x})$, which is likely to be multi-modal. The original formulation of the GAN objective function \cite{goodfellow2014} has been shown to lead to unimodal pdfs, even if the training set pdf itself is multi-modal \cite{goodfellowtutorial}. The behavior of a generator to represent multi-modal pdfs by a pdf with fewer modes is called mode-collapse \cite{goodfellowtutorial}. This behavior may occur due to the fact that there is no incentive for diversity in GAN training. 
 \begin{figure}\centering
  \includegraphics[keepaspectratio=True, width=\textwidth]{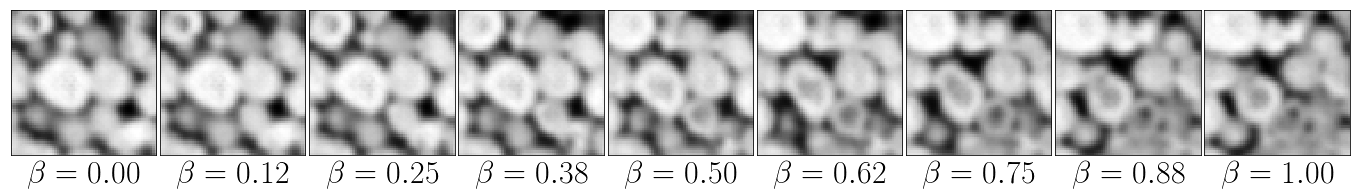}
\caption{Interpolation in the latent space $\mathbf{z}$ performed for the evaluated generator $G_{\mathbf{\theta}}$ shows a smooth interpolation between the start latent vector $\mathbf{z}_{start}$ ($\beta=1$) and the end point $\mathbf{z}_{end}$ ($\beta=0$). An example feature of this can be seen by a bright calcite grain being present in the left most image and slowly being transformed into a spherical grain with significant micro-porosity.}
\label{fig:generator_interpolation}
\end{figure}
Visually the images generated by the presented GAN model are nearly indistinguishable from their real counterparts (Fig.~\ref{fig:comparison_data}). Minkowski functionals and statistical parameters allow us to perform an evaluation of the reconstruction quality. Nevertheless, this does not rule out the fact that the generator may be memorizing the training set, show mode-collapse behavior or output a low diversity of synthetic samples. A generator showing one or more of these behaviors will falsely indicate low errors in the Minkowski functionals, statistical and effective properties.

By visual inspection of the validation set generated by the GAN model, no evidence of identical or repeated features in the generated images could be found. Following the approach by Radford et. al \cite{Radford2016} we perform an interpolation between two points in the latent space  $\mathbf{z}$ (Eq.~\ref{z_int}) as evidence for the generator's ability to learn meaningful representations and to show the absence of memorization. We interpolate between two vectors in the latent space as follows:
\begin{subequations}
\label{equ:interpolation}
\begin{eqnarray}
\mathbf{z}_{start}, \mathbf{z}_{end} \in \mathcal{N}(0, 1)^{512\times 1\times 1\times1} \label{z1_z2}, \ \beta \in [0, 1] \\
\mathbf{z}_{inter} = \beta \ \mathbf{z}_{start} + (1-\beta) \ \mathbf{z}_{end} \label{z_int}
\end{eqnarray}
\end{subequations}
\noindent where $\beta$ is a range of numbers from zero to one.

The smooth transition between the starting image $G_{\mathbf{\theta}}(\mathbf{z}_{start})$ and the endpoint $G_{\mathbf{\theta}}(\mathbf{z}_{end})$ shown in Fig.~\ref{fig:generator_interpolation} indicates that the generator has not memorized the training set and learned a lower-dimensional representation $\mathbf{z}$ that results in meaningful features of the pore-grain micro-structure. Definition of GAN training objectives compatible with high-diversity samples showing no mode-collapse and stable training remains an open problem. Che et. al \cite{che2016mode} present a summary of recent advances to counteract mode-collapse, as well as propose a regularization method improve GAN output variety. Reformulations of the GAN training criterion (Eq.~\ref{equ:minmax}) based on the Wasserstein distance show the ability to model multi-modal densities and allow stable training \cite{gulrajani2017improved}.

\begin{figure}\centering
  \includegraphics[keepaspectratio=True, width=\textwidth]{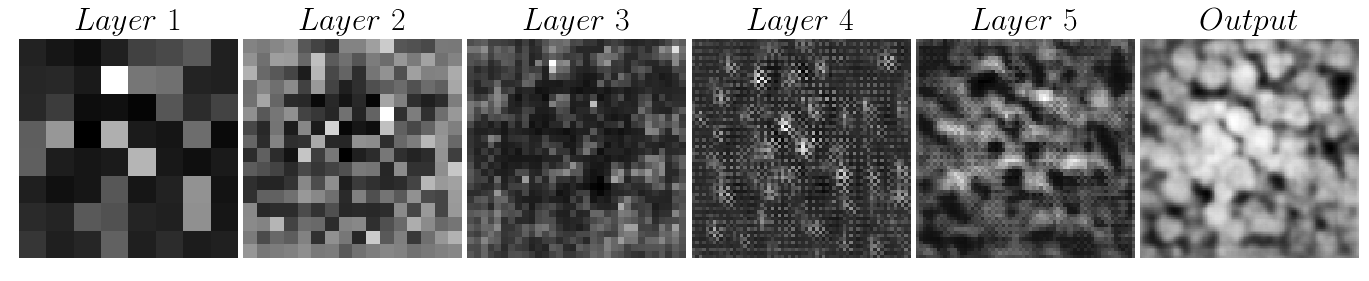}
\caption{Representations of the noise prior $\mathbf{z}$ as it is propagated through the generator $G_{\theta}$. Each layer adds to a multi-scale reconstruction of the final image $G_{\theta}(\mathbf{z})$. The shallow layers 1, 2 and 3 introduce global features of the final image whereas deeper layers add high fidelity details to the output image. Significant noise is still present in layer 3 due to the use of transposed convolution operations, but reduced by the convolution in layer 5.}
\label{fig:layers_output}
\end{figure}

While the input and output to the GAN generator and discriminator is well defined, the interior mechanics of the neural network that result in high-quality reconstructions is not well understood. Rather than treating GANs as a black-box mechanism, it is of interest to evaluate the behavior of the generator and discriminator in more detail. In Fig.~\ref{fig:layers_output} we have extracted the generator's output after each layer's activation function (following the convolution operation and batch normalization). 

Based on the consecutive upsampling of the noise prior $\mathbf{z}$ by each transposed convolution in the generator, we observe a multi-scale feature representation of the final image. Early layers, where the spatial dimensions of the images are small, can be related to global features in the generator output. The final layers create highly detailed representations of the structural features of the reconstructed images. This view of the generator's behavior also helps identify deficiencies in the networks architecture. In layers 3 and 4 we see repeated noise that appears to be following a grid like structure. This is due to the transposed convolutional operation and in parts is diminished by the additional convolution operation prior to the last upsampling operation. This could be alleviated by the use of other convolution based upsampling layers such as the sub-pixel convolution operation \cite{pixelshuffle} or interpolation upsampling (nearest neighbor, bilinear, trilinear).

The discriminator's role is simply to label images as real or "fake", but it also is a critical component in the ability of the generator to learn features in the original image space. The discriminator, in order to distinguish GAN-generated from real training images, needs to learn a unique set of features that distinguish real samples from fake ones. As such, for future work, it may be of interest to use a GAN trained discriminator as a more general classifier or feature representation \cite{arora2017gans}.

\begin{figure}[!b]\centering
  \includegraphics[keepaspectratio=True, width=1.\textwidth]{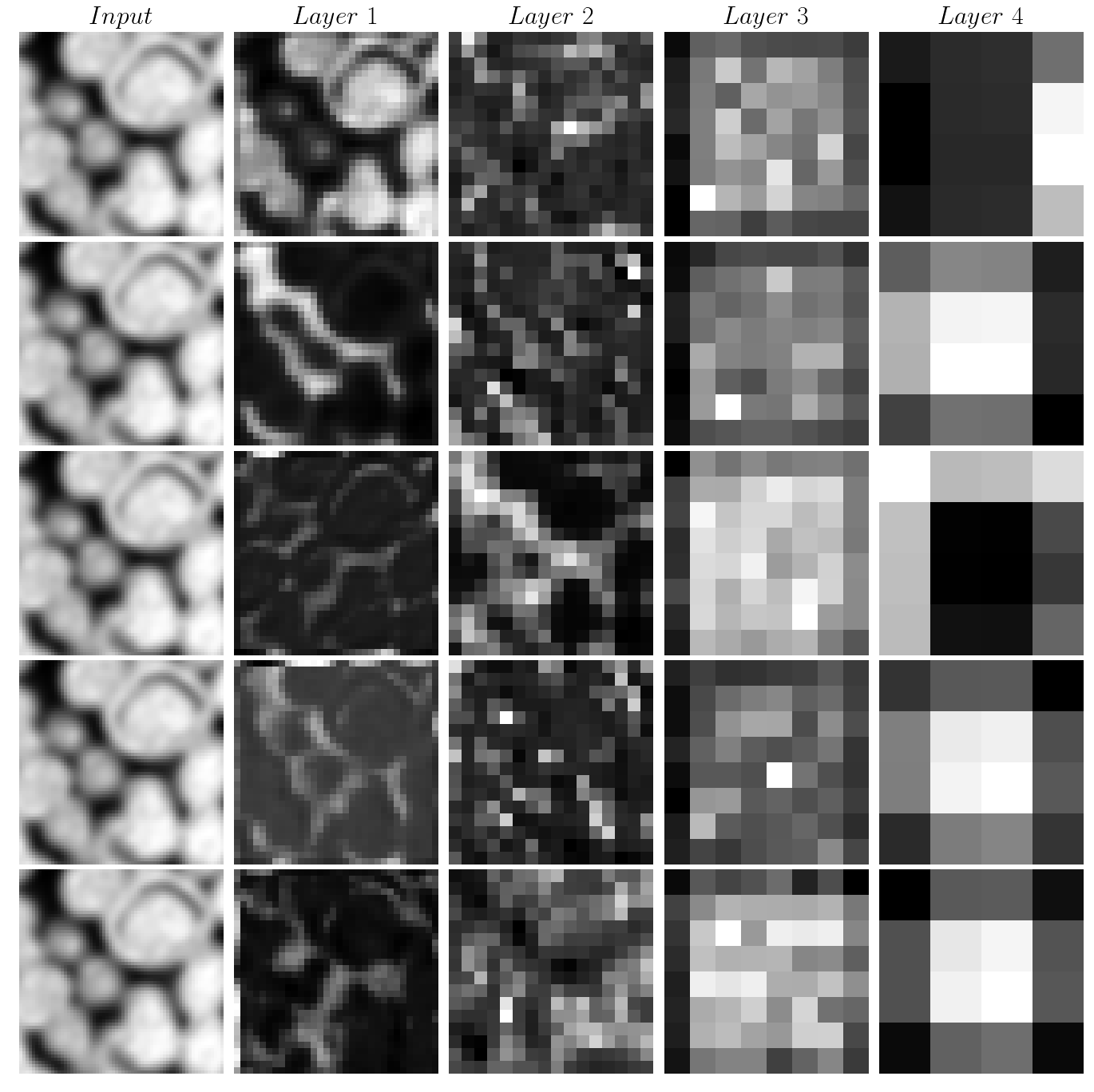}
\caption{An inspection of the behavior of the discriminator's learned feature representations for a training sample of the original Ketton training image. Each column represents one layer of the discriminator network. Each row represents one learned filter kernel in each layer applied to the input (leftmost column).}
\label{fig:features_discriminator}
\end{figure}

Nevertheless, we can perform a similar operation as for the generator and inspect some of the features learned by the discriminator. Figure~\ref{fig:features_discriminator} shows a set of 5 learned filters applied to an image of the Ketton training set. At shallow layers we find that the discriminator has learned to identify the pore space (Layer 1, second row) as well as a number of edge detection filters. Deeper layers in the network represent more abstract representations and after layer 2 no original features of the pore space is distinguishable.

Considering that the samples used to evaluate the statistical and effective properties were not chosen by hand but represent a random sample of generated image based on the GAN model, further improvement can be obtained in the reconstruction results. The discriminator may be used as an evaluation criterion for samples where higher values obtained from the discriminator $D(G_{\theta}(\mathbf{z}))$ indicate that the samples are closer to the real training image dataset. In this way, high-quality reconstructions may be "cherry-picked" by choosing higher scoring representations from a much larger set of reconstructions.

The black-box behavior of GANs as well as their theoretical and practical challenges makes training GAN based generative models difficult. The computational effort and the time necessary to find a set of hyper-parameters that leads to convergence of GAN training is less important if the intent is to create many synthetic realizations of a micro-structure and as such lends itself for applications where a large number of samples are required such as for uncertainty quantification \cite{2017arXiv170801810C,2017arXiv170804975L}.

\section{Conclusion}
\label{sec:conclusion}
We have presented a method to reconstruct micro-structures of porous media based on gray-scale image representations of volumetric porous media. By creating a GAN based model of an oolitic Ketton limestone, we have shown that GANs can learn to represent the statistical and effective properties of segmented representations of the pore space as well as their Minkowski functionals as a function of the image gray-level. In addition to the effective permeability which is associated with a global average of the velocity field, we show that the pore-scale velocity statistical distributions have been recovered by the synthetic GAN based models. We highlight the roles of the discriminator and generator function of the GAN and show that the GAN learns a multi-scale representation of the pore-space based on inference from a latent noise prior. Large hyper-parameter search involved in the deep neural network architectures and learning instabilities make the training of GANs difficult. The high computational cost involved in training GANs is made good use of for applications when very large or many stochastic reconstructions are required. Future work will focus on creating GAN-based methodologies that ensure a valid representation of the underlying data distribution allowing application of GANs for uncertainty quantification of effective material properties.
\newpage
\subsection*{Acknowledgments}
The authors would like to thank Hannah Menke (Imperial College London) for providing the Ketton image dataset. The authors thank H.J. Vogel of UFZ - Helmhotz Center for Environmental Research for making the software library available for public use. O. Dubrule would like to thank Total S.A. for seconding him as a visiting professor at Imperial College London.
\FloatBarrier
\bibliographystyle{unsrt} 
\bibliography{paper}

\end{document}